\documentclass[10pt,twocolumn,letterpaper]{article}

\usepackage{cvpr}
\usepackage{times}
\usepackage{epsfig}
\usepackage{graphicx}
\usepackage{amsmath}
\usepackage{amssymb}
\usepackage[table]{xcolor} 
\usepackage{subcaption}
\usepackage{enumitem}

\usepackage[breaklinks=true,bookmarks=false]{hyperref}

\cvprfinalcopy % *** Uncomment this line for the final submission

\def\cvprPaperID{****} % *** Enter the CVPR Paper ID here

\usepackage{color}

\ifcvprfinal\fi
\begin{document}
%%%%%%%%% TITLE
\title{Rich Image Captioning in the Wild}

\author{Kenneth Tran, Xiaodong He, Lei Zhang, Jian Sun\\
	Cornelia Carapcea, Chris Thrasher, Chris Buehler, Chris Sienkiewicz\\\\
Microsoft Research\\\\
{\tt\small \{ktran,xiaohe\}@microsoft.com}\thanks{Corresponding authors}
% For a paper whose authors are all at the same institution,
% omit the following lines up until the closing ``}''.
% Additional authors and addresses can be added with ``\and'',
% just like the second author.
% To save space, use either the email address or home page, not both
}

\maketitle
%\thispagestyle{empty}

%%%%%%%%% ABSTRACT
\begin{abstract}
We present an image caption system that addresses new challenges of automatically describing images in the wild. The challenges include high quality caption quality with respect to human judgments, out-of-domain data handling, and low latency required in many applications. Built on top of a  state-of-the-art framework, we developed a deep vision model that detects a broad range of visual concepts, an entity recognition model that identifies celebrities and landmarks, and a confidence model for the caption output.  Experimental results show that our caption engine outperforms previous state-of-the-art systems significantly on both in-domain dataset (i.e. MS COCO) and out-of-domain datasets. %Anonymous We also make the system publicly accessible as a part of the Microsoft Cognitive Services.
\end{abstract}

%%%%%%%%% BODY TEXT
\section{Introduction}
\begin{figure}
	\centering
	\includegraphics[width=0.7\linewidth]{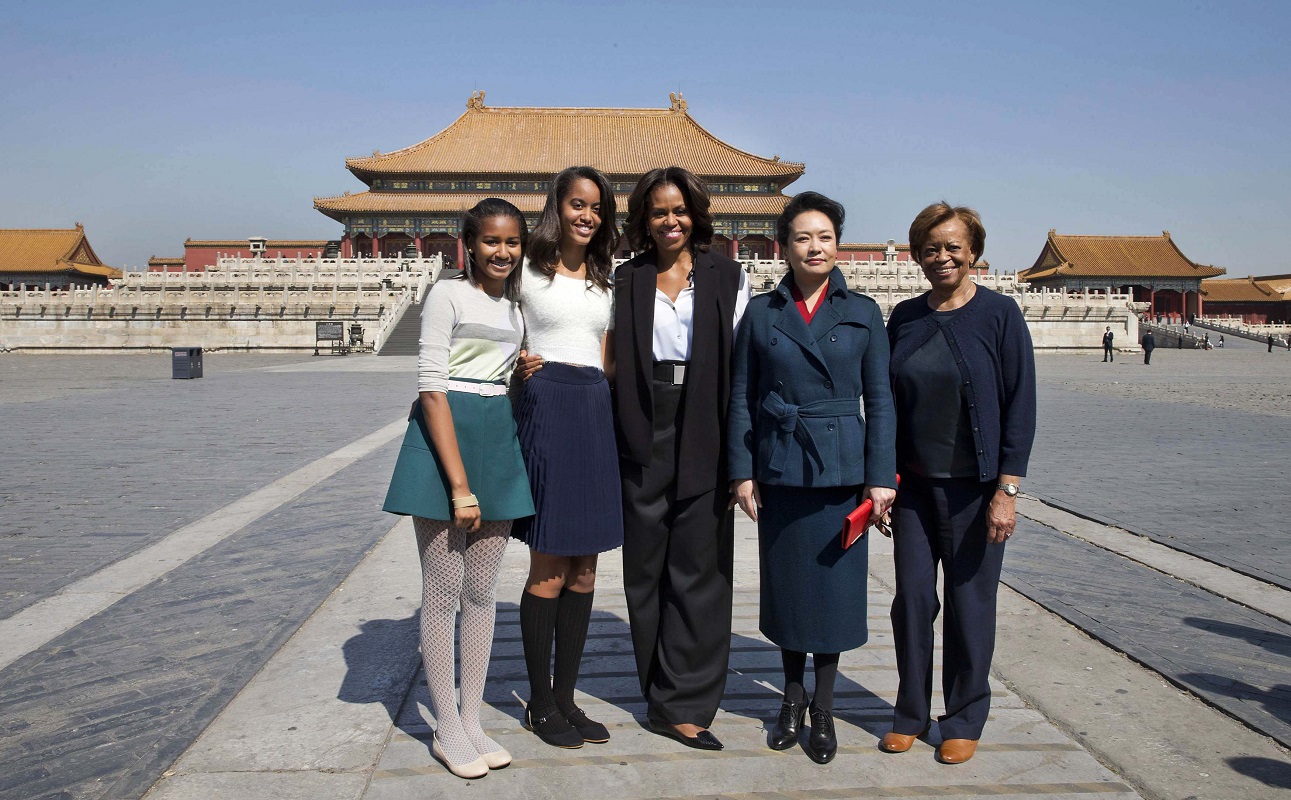}\\
	\small{\emph{``Sasha Obama, Malia Obama, Michelle Obama, Peng Liyuan et al. posing for a picture with Forbidden City in the background."}}\\
 	\vspace{1em}
	\includegraphics[width=0.7\linewidth]{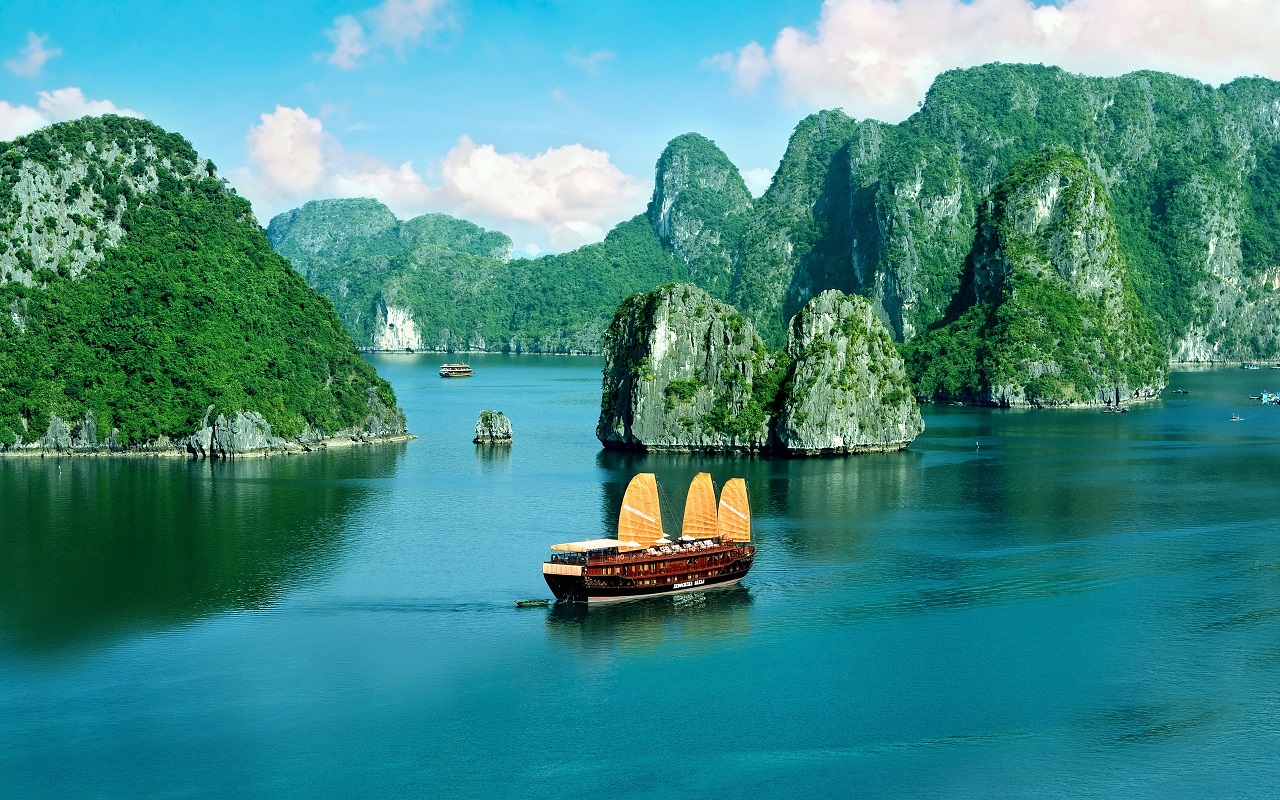}\\
	\small{\emph{``A small boat in Ha-Long Bay."}}
	\caption{Rich captions enabled by entity recognition}
	\label{fig:showcase}
\end{figure}

Image captioning is a fundamental task in Artificial Intelligence which describes objects, attributes, and relationship in an image, in a natural language form. It has many applications such as semantic image search, bringing visual intelligence to chatbots, or helping visually-impaired people to see the world around them. Recently, image captioning has received much interest from the research community (see \cite{vinyals2015show,xu2015show,you2016caption,donahue2015long,fang2015captions,karpathy2015deep,johnson2015densecap}). 

The leading approaches can be categorized into two streams. One stream takes an end-to-end, \emph{encoder-decoder} framework adopted from machine translation. For instance, \cite{vinyals2015show} used a CNN to extract high level image
features and then fed them into a LSTM to generate caption. \cite{xu2015show} went one step further by introducing the attention mechanism. The other stream applies a \emph{compositional} framework. For example, \cite{fang2015captions} divided the caption generation into several parts: word detector by a CNN, caption candidates generation by a maximum entropy model, and sentence re-ranking by a deep multimodal semantic model.

\begin{figure*}
	\includegraphics[width=\textwidth]{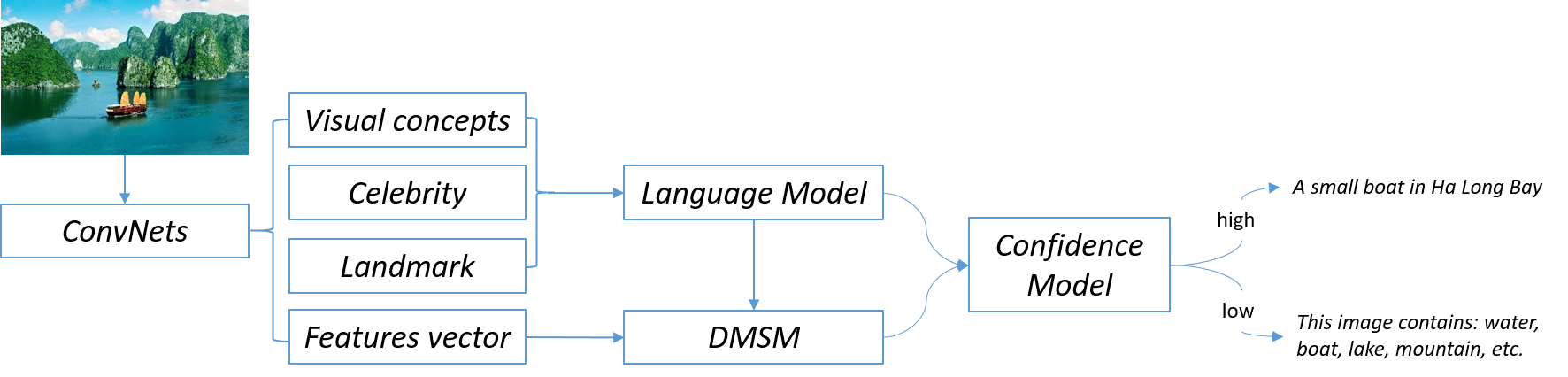}
	\caption{Illustration of our image caption pipeline.}
	\label{fig:pipeline}
\end{figure*}

However, while significant progress have been reported \cite{you2016caption,vinyals2015show,donahue2015long,fang2015captions}, most of the systems in literature are evaluated on academic benchmarks, where the experiments are based on test images collected under a controlled environment which have similar distribution to the training examples. It is unclear how these systems perform on open-domain images.

Furthermore, most of the image captioning systems only describe generic visual content without identifying key entities. The entities, such as celebrities and landmarks, are important pieces in our common sense and knowledge. In many situations (e.g., Figure~\ref{fig:showcase}), the entities are the key information in an image.

In addition, most of the literature report results in automatic metrics such as BLEU \cite{papineni2002bleu}, METEOR \cite{agarwal2008meteor}, and CIDEr \cite{vedantam2015cider}. Although these metrics are handy for fast development and tuning, there exists a substantial discrepancy between these metrics and human's judgment \cite{devlin2015language,kulkarni2011baby,callison2006re}. Their correlation to human’s judgment could be even weaker when evaluating captions with entity information integrated. 

In this paper, we present a captioning system for open domain images. We take a compositional approach by starting from one of the state-of-the-art image captioning framework \cite{fang2015captions}. To address the challenges when describing images in the wild, we enriched the visual model by detecting a boarder range of visual concepts and recognizing celebrities and landmarks for caption generation (see examples in Figure \ref{fig:showcase}). Further, in order to provide graceful handling for images that are difficult to describe, we built a confidence model to estimate a confidence score for the caption output based on the vision and text features, and provide a back-off caption for these difficult cases. We also developed an efficient engine that integrates these components and generates the caption within one second end-to-end on a 4-core CPU. 

In order to measure the quality of the caption from the human’s perspective, we carried out a series of human evaluations through crowd souring, and report results based on human's judgments. Our experimental results show that the proposed system outperforms a previous state-of-the-art system \cite{fang2015captions} significantly on both in-domain dataset (MS COCO \cite{lin2014microsoft}), and out-of-domain datasets (Adobe-MIT FiveK \cite{fivek} and a dataset consisting randomly sampled images from Instagram \footnote{Instagram data: https://gist.github.com/Anonymous%zer0n/061d6c5e0cb80b56d0a3
}.) Notably, we improved the human satisfaction rate by 94.9\% relatively on the most challenging Instagram dataset.

\section{Model architecture}
Following Fang et al. \cite{fang2015captions}, we decomposed the image caption system into independent components, which are trained separately and integrated in the main pipeline. The main components include
\begin{itemize}[noitemsep,topsep=0pt]
	\item a deep residual network-based vision model that detects a broad range of visual concepts,
	\item a language model for candidates generation and a deep multimodal semantic model for caption ranking,
	\item an entity recognition model that identifies celebrities and landmarks,
	\item and a classifier for estimating the confidence score for each output caption.
\end{itemize}
Figure \ref*{fig:pipeline} gives an overview of our image captioning system.

\subsection{Vision model using deep residual network}
\label{sec:resnet}
\newcommand{\ve}[1]{\mathbf{#1}} % for displaying a vector
\newcommand{\ma}[1]{\mathrm{#1}} % for displaying a matrix

\begin{figure}
\centering
\includegraphics[width=0.4\linewidth]{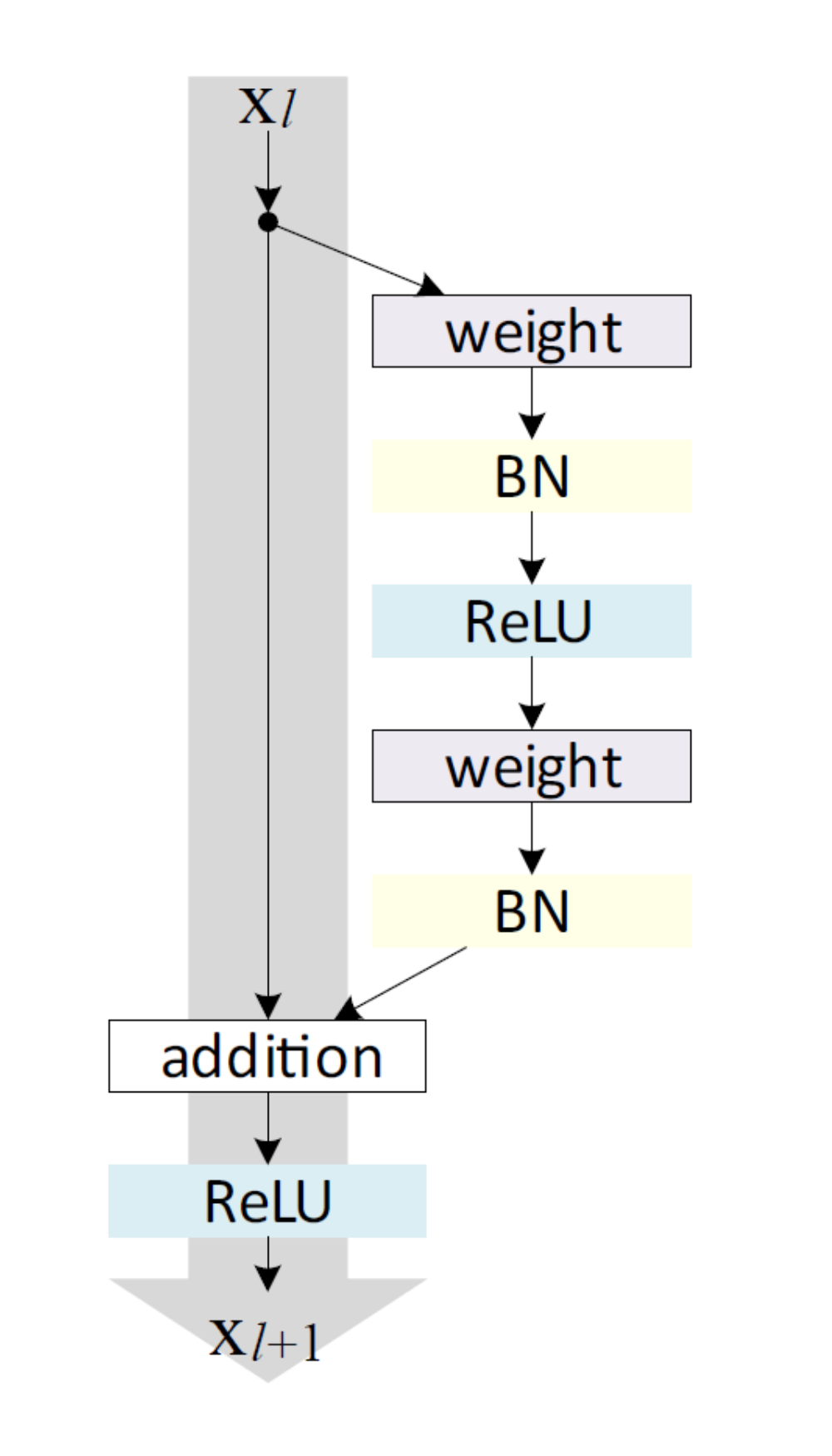}
\caption{A residual unit. Here $\ve{x}_{l}$/$\ve{x}_{l+1}$ is the input/output feature to the $l$-th Residual Unit. Weight, BN, ReLU are linear convolution, batch normalization~\cite{Ioffe2015}, and Rectified Linear Unit~\cite{Nair2010} layers. }
\label{fig:ressidual_unit}
\end{figure}

Deep residual networks (ResNets) \cite{he2016resnet} consist of many stacked ``Residual Units". Each residual unit (Fig. \ref{fig:ressidual_unit}) can be expressed in a general form:
\begin{gather}
	\vspace{-.4em}
	\ve{y}_{l} = h(\ve{x}_{l}) + \mathcal{F}(\ve{x}_{l}, \mathcal{W}_l), \nonumber\\
	\ve{x}_{l+1} = f(\ve{y}_{l}) \nonumber,
	\vspace{-.4em}
\end{gather}
where $\ve{x}_{l}$ and $\ve{x}_{l+1}$ are input and output of the $l$-th unit, and $\mathcal{F}$ is a residual function. In \cite{he2016resnet}, $h(\ve{x}_{l}) = \ve{x}_{l}$ is an identify mapping and $f$ is a ReLU \cite{Nair2010} function.
ResNets that are over 100-layer deep have shown state-of-the-art accuracy for several challenging recognition tasks on ImageNet \cite{ILSVRC15} and MS COCO \cite{Lin2014} competitions. The central idea of ResNets is to learn the additive residual function $\mathcal{F}$ with respect to $h(\ve{x}_{l})$, with a key choice of using an identity mapping $h(\ve{x}_{l}) = \ve{x}_{l}$. This is realized by attaching an identity skip connection (``shortcut'').

\paragraph{Training.} In order to address the open domain challenge, we trained two classifiers. The first classifier was trained on MS COCO training data, for 700 visual concepts. And the second one was trained on an image set crawled from commercial image search engines, for 1.5K visual objects. The training started from a 50-layer ResNet, pre-trained on ImageNet 1K benchmark. To handle multiple-label classification, we use sigmod output layer without softmax normalization.

\paragraph{Testing.} To make the testing efficient, we apply all convolution layers on the input image once to get a feature map (typically non-square) and perform average pooling and sigmoid output layers. Not only our network provides more accurate predictions than VGG \cite{simonyan2014very}, which is used in many caption systems \cite{fang2015captions,xu2015show,karpathy2015deep}, it is also order of magnitude faster. The typical runtime of our ResNet is 200ms on a desktop CPU (single core only).

\subsection{Language and semantic ranking model}

Unlike many recent works \cite{vinyals2015show,xu2015show,karpathy2015deep} that use LSTM/GRU (so called gated recurrent neural network or GRNN) for caption generation, we follow \cite{fang2015captions} to use a maximum entropy language model (MELM) together with a deep multimodal similarity model (DMSM) in our caption pipeline. While MELM does not perform as well as GRNN in terms of perplexity, this disadvantage is remedied by DMSM. Devlin et al. \cite{devlin2015language} shows that while MELM+DMSM gives the same BLEU score as GRNN, it performs significantly better than GRNN in terms of human judgment. The results from the MS COCO 2015 captioning challenge\footnote{http://mscoco.org/dataset/\#captions-leaderboard} also show that the MELM+DMSM based entry \cite{fang2015captions} gives top performance in the official human judgment, tying with another entry using LSTM. 

\begin{figure}
	\centering
	\includegraphics[width=0.3\textwidth]{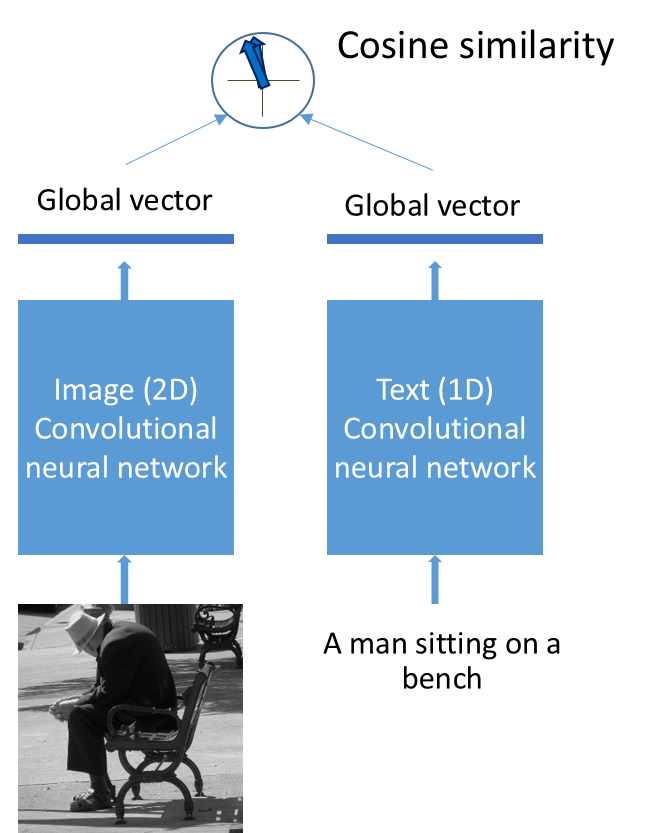}
	\caption{Illustration of deep multimodal semantic model}
	\label{fig:dmsm}
\end{figure}

In the MELM+DMSM based framework, the MELM is used together with beam search as a candidate caption generator. Similar to the text-only deep structured semantic model (DSSM) \cite{huangdssm,shencdssm}, The DMSM is illustrated in Figure \ref{fig:dmsm}, which consists of a pair of neural networks, one for mapping each input modality to a common semantic space. These two neural networks are trained jointly\cite{fang2015captions}. In training, the data consists of a set of image/caption pairs. The loss function minimized during training represents the negative log posterior probability of the caption given the corresponding image. The image model reuses the last pooling layer extracted in the word detection model, as described in section \ref{sec:resnet}, as feature vector and stacks one more fully-connected layer with Tanh non-linearity on top of this representation to obtain a final representation of the same size as the last layer of the text model. We learn the parameters in this additional layer during DMSM training. The text model is based on a one-dimensional convolutional neural network similar to  \cite{shencdssm}. The DMSM similarity score is used as the main signal for ranking the captions, together with other signals including language model score, caption length, number of detected words covered in the caption, etc. 

In our system, the dimension is set to be 1000 for the global vision vector and the global text vector, respectively. The MELM and the DMSM are both trained on the MS COCO dataset \cite{lin2014microsoft}. Similar to \cite{huangdssm}, character-level word hashing is used to reduce the dimension of the vocabulary.

\subsection{Celebrity and landmark recognition}

The breakthrough in deep learning makes it possible to recognize visual entities such as celebrities and landmarks and link the recognition result to a knowledge base such as Freebase \cite{bollacker2008freebase}. We believe providing entity-level recognition results in image captions will bring valuable information to end users. 

The key challenge to develop a good entity recognition model with wide coverage is collecting high quality training data. To address this problem, we followed and generalized the idea presented in \cite{zhang2012finding} which leverages duplicate image detection and name list matching to collect celebrity images. In particular, we ground the entity recognition problem on a knowledge base, which brings in several advantages. First, each entity in a knowledge base is unique and clearly defined without unambiguity, making it possible to develop a large scale entity recognition system. Second, each entity normally has multiple properties (e.g. gender, occupation for people, and location, longitude/latitude for landmark), providing rich and valuable information for data collecting, cleaning, multi-task learning, and image description.

We started with a text-based approach similar to \cite{zhang2012finding} but using entities that are catalogued in the knowledge base rather than celebrity names for high precision image and entity matching. To further enlarge the coverage, we also scrape commercial image search engines for more entities and check the consistency of faces in the search result to remove outliers or discard those entities with too many outlier faces. After these two stages, we ended up with a large-scale face image dataset for a large set of celebrities.

%Recognizing such a large number of 200K celebrities is another challenge. A typical approach for this problem is $k$-nearest neighbor search-based. That is, for any query face image, we can search for its $k$-nearest neighbors to get its prediction result. However, this approach is quite sensitive to data noise and normally has a poor generalization ability, not to say the computational cost is proportional to the number of images in the dataset. 
%Directly training a CNN on $200$K classes from scratch may have troubles on the convergence. Instead, we started a model on 1K celebrities, each of which has a sufficient number of face images. Then we fine-tuned the model on $200$K celebrities. The whole training process follows the standard setting as described in \cite{krizhevsky2012imagenet}. We decreased the learning rate from $0.01$ to $0.001$ and $0.0001$, at $100$K and $200$K iterations. We use a threshold of $0.9$ to strike a good balance between high precision ($99\%$ on a validation set) and good recall.

\begin{figure}
	\centering
	\includegraphics[width=\linewidth]{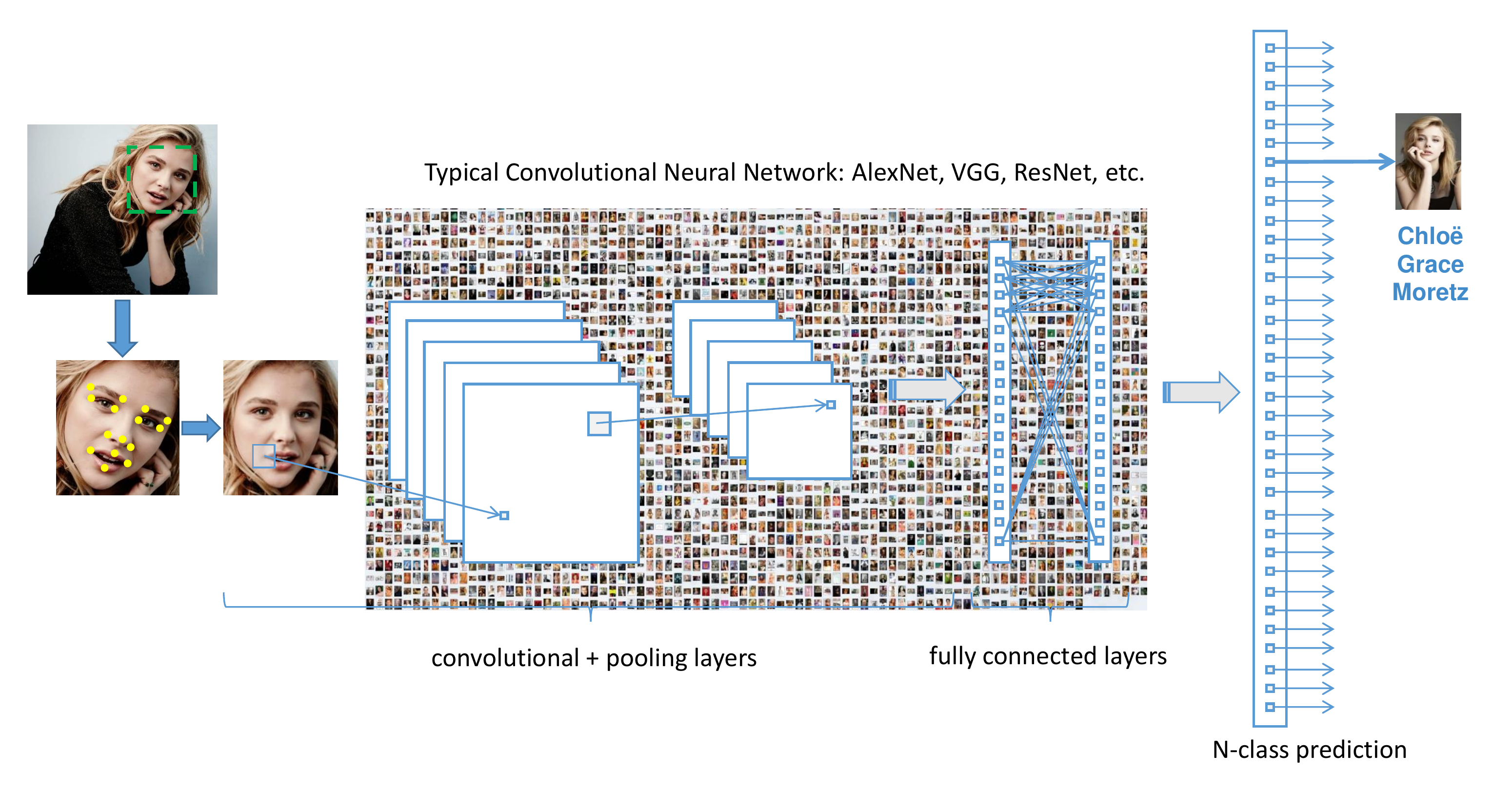}\\
	\caption{Illustration of deep neural network-based large-scale celebrity recognition}
	\label{fig:celeb_recognition}
\end{figure}

To recognize a large set of celebrities, we resorted to deep convolutional neural network (CNN) to learn an extreme classification model, as shown in Figure \ref{fig:celeb_recognition}. Training a network for a large set of classes is not a trivial task. It is hard to see the model converge even after a long run due to the large number of categories. To address this problem, we started from training a small model using AlexNet \cite{krizhevsky2012imagenet} for $500$ celebrities, each of which has a sufficient number of face images. Then we used this pre-trained model to initialize the full model of a large set of celebrities. The whole training process follows the standard setting as described in \cite{krizhevsky2012imagenet}. After the training is finished, we use the final model to predict celebrities in images by setting a high threshold for the final softmax layer output to ensure a high precision celebrity recognition rate. 

%Preliminary experiments show that a threshold of $0.9$ can ensure a prediction precision of $99\%$ with a decent coverage on out of training data images, e.g. images from Instagram.

We applied a similar process for landmark recognition. One key difference is that it is not straightforward to identify a list of landmarks that are visually recognizable although it is easy to get a list of landmarks or attractions from a knowledge base. This implies that data collection and visual model learning are two closely coupled problems. To address this challenge, we took an iterative approach. That is, we first collected a training set for about 10K landmarks selected from a knowledge base to train a CNN model for 10K landmarks. Then we leveraged a validation dataset to evaluate whether an landmark is visually recognizable, and remove from the training set those landmarks which have very low prediction accuracy. After several iterations of data cleaning and visual model learning, we ended up with a model for about 5K landmarks.

\subsection{Confidence estimation}
We developed a logistic regression model to estimate a confidence score for the caption output. The input features include the DMSM's vision and caption vectors, each of size 1000, coupled with the language model score, the length of the caption, the length-normalized language model score, the logarithm of the number of tags covered in the caption, and the DMSM score.

The confidence model is trained on  2.5K image-caption pairs, with human labels on the quality (\emph{excellent, good, bad, embarrassing}). The images used in the training data is a mix of 750 COCO, 750 MIT, and 950 Instagram images in a held-out set. 

\section{Evaluation}
We conducted a series of human evaluation experiments through CrowdFlower, a crowd sourcing platform with good quality control\footnote{http://www.crowdflower.com/}. The human evaluation experiments are set up such as for each pair of image and generated caption, the caption is rated on a 4-point scale: \textit{Excellent}, \textit{Good}, \textit{Bad}, or \textit{Embarrassing} by three different judges. In the evaluation, we specify for the Judges that \textit{Excellent} means that the caption contains all of the important details presented in the picture; \textit{Good} means that the caption contains some instead of all the important details presented in the picture and no errors; \textit{Bad} means the caption may be misleading (e.g., contains errors, or miss the gist of the image); and \textit{Embarrassing} means that the caption is totally wrong, or may upset the owner or subject of the image.

In order to evaluate the captioning performance for images in the wild, we created a dataset from Instagram. Specifically, we collected 100 popular Instagram accounts on the web, and for each account we constructed a query with the account name plus ``instagram", e.g. ``iamdiddy instagram", to scrape the top 100 images from Bing image search. And finally we obtained a dataset of about 10K images from Instagram, with a wide range of coverage on personal photos. About 12.5\% of images in this Instagram set contain entities that are recognizable by our entity recognition model (mostly are celebrities). Meanwhile, we also reported results on 1000 random samples of the COCO validation set and 1000 random samples of the MIT test set, Since the MELM and the DMSM are both trained on the COCO training set, the results on the COCO test set and the MIT test set represent the performance on in-domain images and out-of-domain images, respectively.

We communicated with the authors of Fang et al. \cite{fang2015captions}, one of the two winners of the MS COCO 2015 Captioning Challenge, to obtain the caption output of our test images from their system. For our system, we evaluated three different settings: \textit{Basic} with no confidence thresholding and no entity recognition, \textit{Basic+Confi.} with confidence thresholding but no entity recognition, and \textit{Full} with both confidence thresholding and entity recognition on. For \textit{Basic+Confi.} and \textit{Full}, we use templates such as \textit{``this image is about \$\{top visual concept\}"}, or \textit{``a picture of \$\{entity\}"} if entity recognizer fires, instead of the caption generated by the language model, whenever the confidence score is below 0.25. The results are presented in Tables \ref*{t:eval_coco}, \ref*{t:eval_mit}, and \ref*{t:eval_instagram}. Since the COCO and MIT images were collected in such a way that does not surface entites, we do not report \textit{Full} in Tables \ref{t:eval_coco} and \ref{t:eval_mit}.

As shown in the results, we have significantly improved the performance over a previous state-of-the-art system in terms of human evaluation. Specifically, the in-domain evaluation results as reported in Table \ref*{t:eval_coco} show that, compared to the baseline by Fang et al., our \textit{Basic} system reduces the \textit{Bad} and \textit{Embarrassing} rates combined by 6.0\%. Moreover, our system significantly improves the portion of captions that are rated as \textit{Excellent} by more than 10\%, mainly thanks to the deep residual network based vision model, plus refinement of the parameters of the engine and other components. Integrating confidence classifier to the system helps reduce the \textit{Bad} and \textit{Embarrassing} rates further. 

The results on the out-of-domain MIT test set are presented in Table \ref*{t:eval_mit}. We observed similar degree of improvements by using the new vision model. More interestingly, the confidence classifier helps significantly on this dataset. E.g., the rate of \textit{Satisfaction}, a combination of \textit{Excellent} and \textit{Good}, is further improved by more than 10\%. 

\begin{figure*}
	%\centering
	\includegraphics[width=0.95\textwidth]{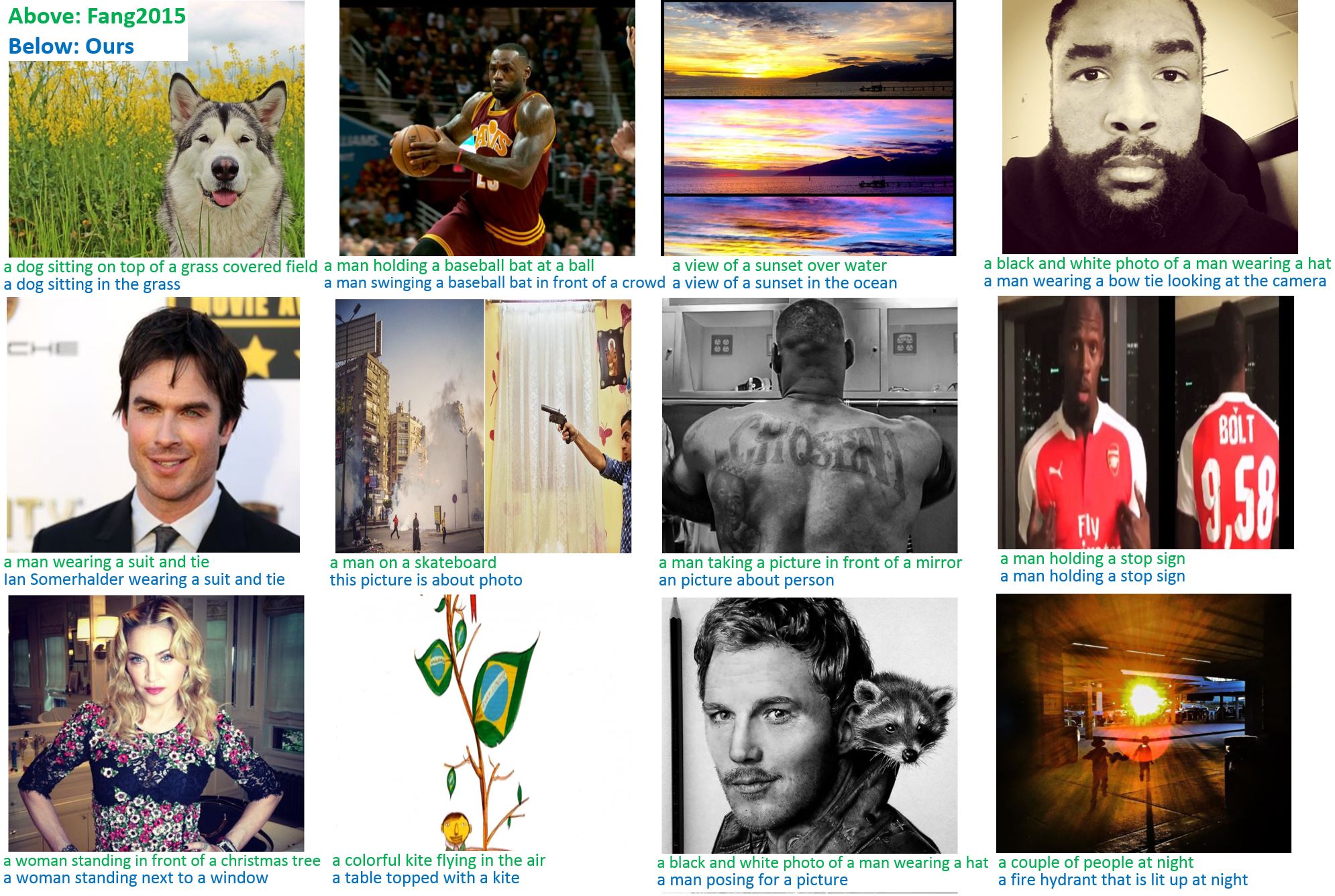}\\
	\includegraphics[width=0.95\textwidth]{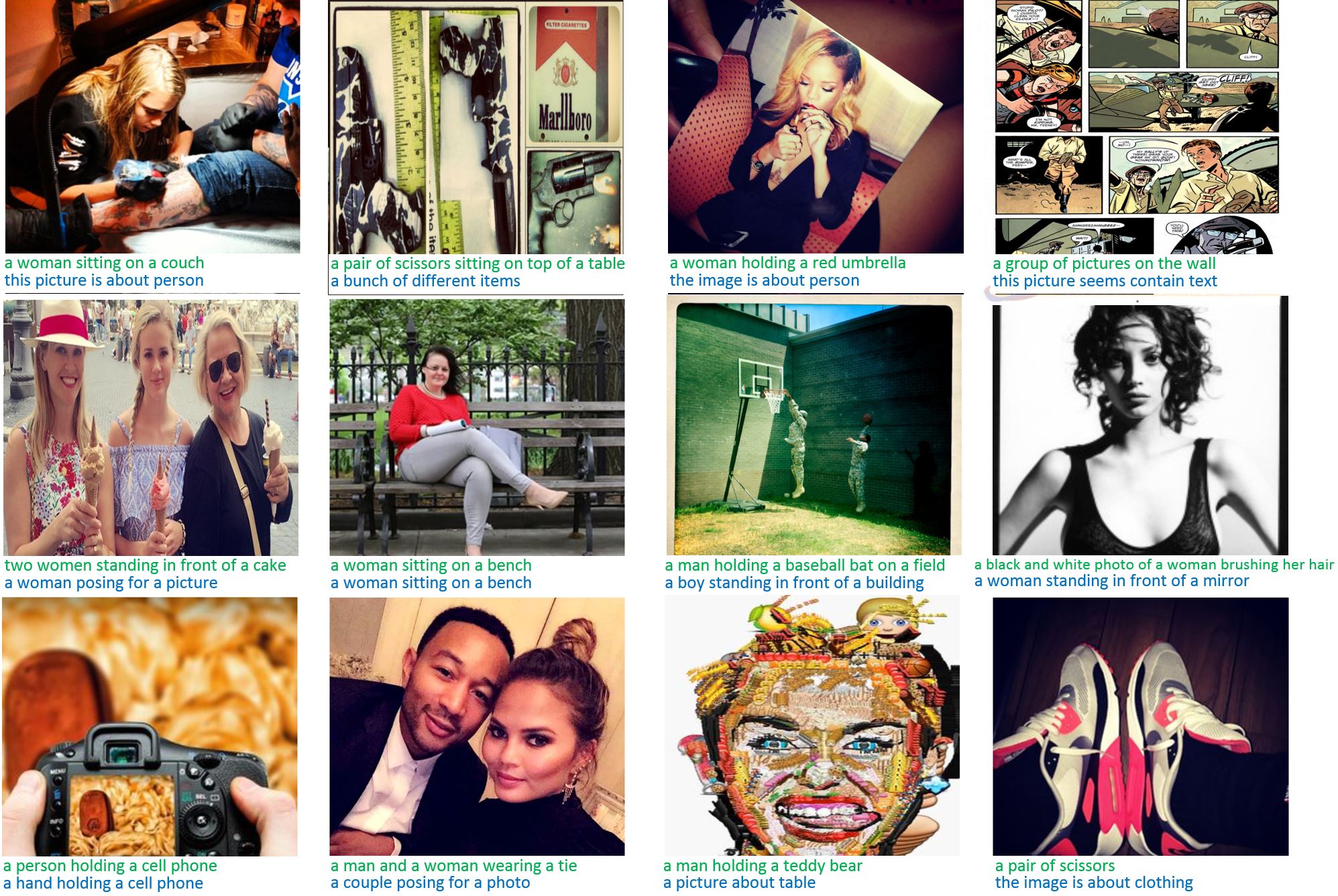}
	\caption{Qualitative results of images  randomly sampled from the Instagram test set, with Fang2015 caption in green (above) and our system's caption in blue (below) for each image.}
	\label{fig:examples}
\end{figure*}

Instagram data set contains many images that are filtered images or handcrafted abstract pictures, which are difficult for the current caption system to process (see examples in Figure \ref{fig:examples}). In the Instagram domain, the results in Table \ref*{t:eval_instagram} shows that both the baseline and our \textit{Basic} system perform quite poorly, scoring a \textit{Satisfaction} rate of 25.4\% and 31.5\%, respectively. However, by integrating confidence classifier in the system, we improve the \textit{Satisfaction} rate to 47.9\%. The \textit{Satisfaction} rate is further improved to 49.5\% after integrating the entity recognition model, representing a 94.9\% relative improvement over the baseline. In Figure \ref{fig:examples}, we show a bunch of images randomly sampled from the Instagram test set. For each image, we also show the captions generated by the baseline system (above, in green) and our \textit{Full} system (below, in blue), respectively.

We further investigated the distribution of confidence scores in each of the \textit{Excellent}, \textit{Good}, \textit{Bad}, and \textit{Embarrassing} category on the Instagram test set using the \textit{Basic} setting. The means and the standard deviations are reported in Table \ref*{t:conf_instagram}. We observed that in general the confidence scores align with the human judgements well. Therefore, based on the confidence score, more sophisticated solutions could be developed to handle difficult images and achieve a better user experience.

We also want to point out that, integrating the entity in the caption greatly improves the user experience, which might not be fully reflected in the 4-point rating. For example, for the first image in the second row of Figure \ref{fig:examples}, the baseline gives a caption ``\textit{a man wearing a suit and tie}", while our system produces ``\textit{Ian Somerhalder wearing a suit and tie}" thanks to the entity recognition model. Although both caption outputs are rated as \textit{Excellent}, the latter provides much richer information than the baseline.

\begin{table}
	\centering
	\begin{tabular}{l*{4}{l}}
		\hline
		\rowcolor{gray!20}
		System & Excel & Good & Bad & Emb \\
		\hline
		Fang et al. & 40.6\% & 26.8\% & 28.8\% & 3.8\% \\
		Ours (Basic) & 51.4\% & 22.0\% & 23.6\% & 3.0\% \\
		Ours (Basic+Confi.) & 51.8\% & 23.4\% & 22.5\% & 2.3\% \\
		\hline
	\end{tabular}
	\caption{Human evaluation on 1K random samples of the COCO val-test set}
	\label{t:eval_coco}
\end{table}

\begin{table}
	\centering
	\begin{tabular}{l*{4}{l}}
		\hline
		\rowcolor{gray!20}
		System & Excel & Good & Bad & Emb \\
		\hline
		Fang et al. & 17.8\% & 18.5\% & 55.8\% & 7.9\% \\
		Ours (Basic) & 23.9\% & 21.0\% & 49.0\% & 6.1\% \\
		Ours (Basic+Confi.) & 28.2\% & 27.5\% & 39.3\% & 5.0\% \\
		\hline
	\end{tabular}
	\caption{Human evaluation on 1K random samples of the MIT test set}
	\label{t:eval_mit}
\end{table}

\begin{table}
	\centering
	\begin{tabular}{l*{4}{l}}
		\hline
		\rowcolor{gray!20}
		System & Excel & Good & Bad & Emb \\
		\hline
		Fang et al. & 12.0\% & 13.4\% & 63.0\% & 11.6\% \\
		Ours (Basic) & 15.1\% & 16.4\% & 60.0\% & 8.4\% \\
		Ours (Basic+Confi.) & 23.3\% & 24.6\% & 47.0\% & 5.1\% \\
		Ours (Full) & 25.4\% & 24.1\% & 45.3\% & 5.2\%\\
		\hline
	\end{tabular}
	\caption{Human evaluation on Instagram test set, which contains 1380 random images from the 10K Instagram images that we scraped.}
	\label{t:eval_instagram}
\end{table}

\begin{table}
	\centering
	\begin{tabular}{l*{4}{l}}
		\hline
		\rowcolor{gray!20}
		  & Excel & Good & Bad & Emb \\
		\hline
		mean & 0.59 & 0.51 & 0.26 & 0.20 \\
		stdev & 0.21 & 0.23 & 0.21 & 0.19 \\
		\hline
	\end{tabular}
	\caption{mean and standard deviation of confidence scores in each category, measured on the Instagram test set under the \textit{Basic} setting.}
	\label{t:conf_instagram}
\end{table}

\section{Conclusion}
This paper presents a new state-of-the-art image caption system with respect to human evaluation. To encourage reproducibility and facilitate further research, we have deployed our system and made it publicly accessible. %Anonymous as a part of Anonymous% the Microsoft Cognitive Services.

%For future work, we plan to extract and integrate richer information, more than just celebrities and landmarks, to the captions. Another equally interesting direction is adding style (e.g. serious or humorous) to the captions.

\section{Acknowledgments}
The authors are grateful to Li Deng, Jacob Devlin, Delong Fu, Ryan Galgon, Jianfeng  Gao, Yandong Guo, Ted Hart, Yuxiao Hu, Ece Kamar, Anirudh Koul, Allison Light, Margaret Mitchell, Yelong Shen, Lucy Vanderwende, and Geoffrey Zweig for valuable discussions. 

{\small
\bibliographystyle{ieee}
\bibliography{references}
}

\end{document}